\documentclass{article}

\usepackage{PRIMEarxiv}

\usepackage[utf8]{inputenc} 
\usepackage[T1]{fontenc}    
\usepackage{hyperref}       
\usepackage{url}            
\usepackage{booktabs}       
\usepackage{amsfonts}       
\usepackage{nicefrac}       
\usepackage{microtype}      
\usepackage{lipsum}
\usepackage{fancyhdr}       
\usepackage{graphicx}       
\usepackage{caption}
\usepackage{amsmath}
\usepackage{amssymb}
\usepackage{ragged2e} 
\usepackage{booktabs,makecell, multirow}
\usepackage{xcolor, colortbl}
\usepackage{cancel}
\graphicspath{{media/}}     

\title{Trajectory Densification and Depth from Perspective-based Blur
}

\author{
  Tianchen Qiu, Qirun Zhang, Jiajian He, Zhengyue Zhuge, Jiahui Xu, Yueting Chen* \\
  College of Optical Science and Engineering\\
  Zhejiang University \\
  Hangzhou, China\\
  \texttt{\{12530010, zhangqirun, hejiajian2, zgzy, xu\_jiahui, chenyt\}@zju.edu.cn} \\
}

\begin{document}
\maketitle

\includegraphics[width=\linewidth]{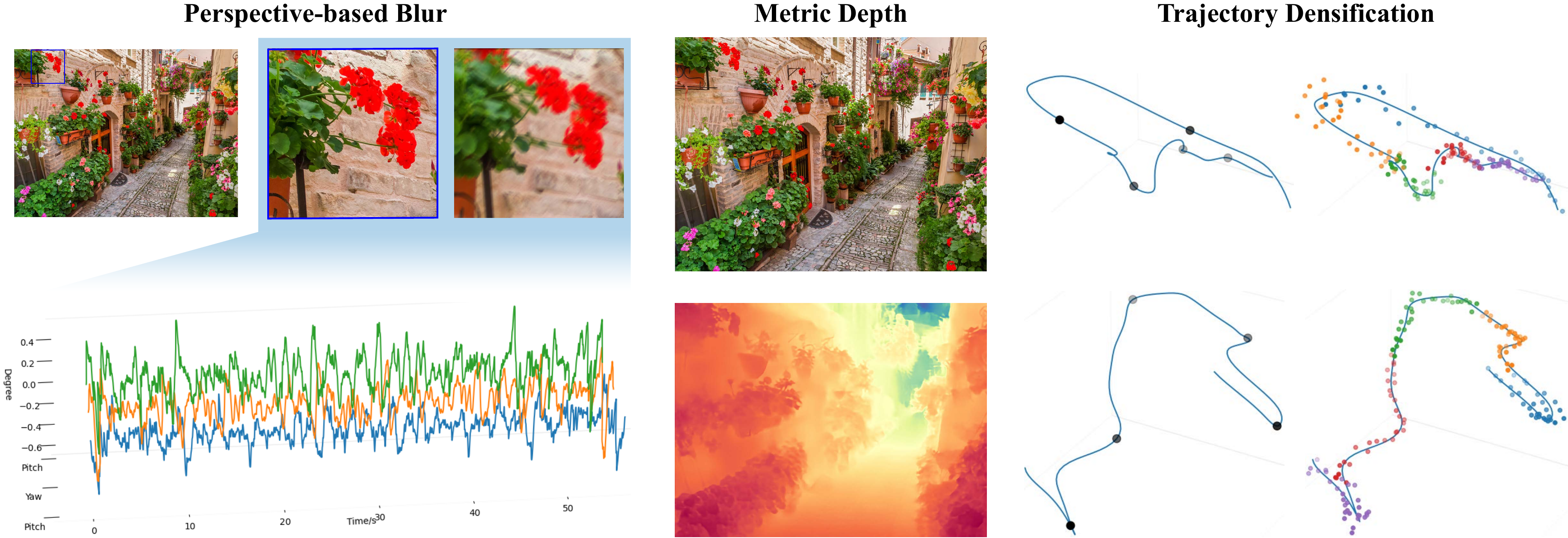}
\vspace{-10pt}
\captionof{figure}{Our method predicts metric depth map and dense trajectory within frames from perspective-based blur, which is caused by camera motion located on the curve of left above. In the trajectory densification part, the left column depicts the computed sparse trajectory, whereas the right column shows the predicted dense one; the blue solid line denotes the ground truth(GT).}
\vspace{5pt}
\label{fig0}

\begin{abstract}
In the absence of a mechanical stabilizer, the camera undergoes inevitable rotational dynamics during capturing, which induces perspective-based blur especially under long-exposure scenarios.
From an optical standpoint, perspective-based blur is depth-position-dependent: objects residing at distinct spatial locations incur different blur levels even under the same imaging settings.
Inspired by this, we propose a novel method that estimate metric depth by examining the blur pattern of a video stream and dense trajectory via joint optical design algorithm.
Specifically, we employ off-the-shelf vision encoder and point tracker to extract video information.
Then, we estimate depth map via windowed embedding and multi-window aggregation, and densify the sparse trajectory from the optical algorithm using a vision–language model.
Evaluations on multiple depth datasets demonstrate that our method attains strong performance over large depth range, while maintaining favorable generalization.
Relative to the real trajectory in handheld shooting settings, our optical algorithm achieves superior precision and the dense reconstruction maintains strong accuracy.
\end{abstract}

\section{Introduction}
\label{sec:intro}

Inferring camera motion from multiple images has been an essential task in computer vision community, for which structure-from-motion (SfM) methods have achieved substantial progress\cite{Multiple_View_Geometry, colmap, dust3r_cvpr24, diffusionsfm}.
That said, most existing approaches emphasize camera position estimation on a per-image basis, while the accompanying view-angle differences are generally considerable.
With the growing adoption of telephoto optical lenses on mobile platforms, imaging without an external stabilizer becomes susceptible to pronounced blur, primarily driven by camera shake.
Furthermore, the in-module optical stabilization system is subject to inherent limitations, such as temporal response lag and residual imaging shift, that introduce perspective-based blur\cite{vibration_simu, simu_res}.
In this regime, pipelines from the SfM literature fall short in producing dense motion trajectory and fail to converge reliably for small-magnitude motions.

From optical standpoints to obtain high-precision sparse trajectory, access to the scene depth map is indispensable.
Depth estimation is a mature line of work that includes metric and relative depth, single image and video streams\cite{zoedepth, depthanything, DepthCrafter, chronodepth}.
Notably, most motion–based approaches target the recovery of relative depth\cite{seurat}; however, for an accurate camera optics model, metric depth is required.
Accordingly, to reconstruct metric depth, they are often integrated with monocular depth prediction methods, producing a dense metric depth map.

Our work propose a novel jointly optical pipeline based on local blur features from monocular video streams, aggregating relative depth information within single temporal window across multiple windows to obtain metric depth map, and condensing trajectory, which is resulted from optics-based algorithm, via vision-language model.
The video stream is encoded with a shared vision Transformer\cite{dinov2}, upon which we attach heterogeneous decoders to derive task-dependent results.
Specifically, considering the vision input of trajectory densification, we route the encoded features through the depth stage’s self-attention layers on a per-frame basis, concatenate them with the terminal depth-decoder predictions.
For training process, we first focus on parameters of depth stage, then train the trajectory stage while freezing the former, only for inference.
Ground-truth depth is sourced from the depth datasets, while the trajectory one corresponds to the input curves used during video simulation\cite{degradation, computational_optics}.

We extensively evaluate our method on diverse datasets, including indoor, outdoor, distant, realistic, and synthetic scenarios.
Both quantitative results of depth estimation and trajectory densification demonstrate that our pipeline achieves state-of-art performance compared with other video depth estimators, outperforming existing SfM methods in terms of the temporal density and accuracy of camera motion.
This main contributions of this work are summarized below:
\begin{itemize}
\item[$\bullet$] We innovate a novel method to reconstruct metric depth map from streaming video sequences, by analyzing the evolution of perspective-based blur across temporal frames, outperforming video depth estimators.
\item[$\bullet$] We propose an optics-based camera trajectory algorithm that leverages a point tracker to obtain delta of predefined points, yielding a frame-by-frame camera trajectory. It exhibits improved accuracy over other methods when evaluated in the small-angle regime.
\item[$\bullet$] We present a two-stage strategy to enable generating dense trajectory, fusing sparsely sampled sequence from first stage and video vision information, in the second stage. It attains strong accuracy when compared against ground truth.
\end{itemize}

\section{Related Work}
\label{sec: Related Work}

\noindent{\bf Structure from motion.} 
SfM aims to jointly recover camera poses and sparse 3D structure from unordered 2D images.
Early theoretical development established the fundamentals of multi-view geometry\cite{longuet-higgins_computer_1981, 3DCV, Multiple_View_Geometry} and bundle adjustment(BA)\cite{BA}.
Classical incremental pipelines, such as Photo Tourism\cite{photo_tourism}, demonstrated Internet-scale 3D reconstruction by sequentially registering views and triangulating 3D points.
Practical SfM systems like VisualSF\cite{visual_sfm} and COLMAP\cite{colmap} combined robust two-view estimation\cite{Fischler1981RandomSC, 8point} with Levenberg–Marquardt(LM) BA\cite{SBA} to deliver highly accurate sparse reconstructions.
Despite their reliability, incremental methods suffer from drift accumulation and are computationally expensive for large-scale datasets.
To address this, global SfM methods estimate all camera orientations and positions in a unified optimization framework.
Representative approaches include 1DSfM\cite{1Dsfm}, which performs robust rotation averaging via pairwise 1D projections, and TheiaSfM\cite{theiasfm}, which jointly refines camera positions using robust cost aggregation.
Hybrid pipelines such as OpenMVG\cite{openMVG} combine global initialization and incremental refinement to ensure global consistency with local precision.

Learning-based SfM has recently emerged to integrate geometric reasoning with deep supervision for better scalability and generalization.
DeepSFM\cite{deepsfm} and BA-Net\cite{Tang2018BANet} embed deep networks into classical BA pipeline. 
DUSt3R\cite{dust3r_cvpr24} directly predicts 3D pointmaps from pairs of input images, demonstrating strong generalization across diverse scenarios.
Extending this framework, MASt3R\cite{mast3r_eccv24} incorporates an additional feature extraction head that enables dense pixel-level correspondence estimation.
With more neural representations\cite{lin2021barf, vggsfm}, geometry-based SfM is increasingly integrated with neural implicit modeling, improving both accuracy and convergence robustness.

\vspace{0.3em}\noindent{\bf Feature point tracking.}
Feature tracking is critical for accurate pose estimation and matching in classical SfM pipelines\cite{active, goodfeature, ijcai}.
Traditional methods relied on optical flow\cite{opticalflow} and KLT trackers\cite{KLTtracker}, while modern dense and sparse descriptors\cite{sift, orb, superpoint} enhanced detection robustness.
Modern works go beyond hand-crafted descriptors, introducing learning-based point-tracking frameworks that explicitly model temporal or spatial coherence.

The Track Any Point (TAP) family of methods\cite{locotracker, tapnet, pip, tapir, cotracker, taptr} improved general-purpose point tracking by constructing dense spatio-temporal correlation volumes.
TAP-Net\cite{tapnet} computed a global cost volume and applied convolutions followed by a soft-argmax for trajectory estimation, while PIPs\cite{pip} iteratively refined track trajectories through MLP-Mixer\cite{mlpmixer} modules.
TAPIR\cite{tapir} integrated TAP-Net with PIPs for better long-term consistency, and CoTracker\cite{cotracker} exploited self-attention in Transformers\cite{attetion} to jointly track multiple interacting points in a unified sequence transformer.
More recent variants such as TAPTR\cite{taptr} and LocoTrack\cite{locotracker} enhance correlation computation through local context aggregation, improving multi-object and non-rigid motion robustness.

\vspace{0.3em}\noindent{\bf Monocular depth estimation.}
Monocular depth estimation (MDE) addresses depth prediction from a single RGB camera.
Classical learning-based MDE approaches, such as MegaDepth\cite{MegaDepthLS} and MiDaS\cite{midas}, demonstrated that large and diverse training datasets significantly improve generalization across domains.
Recent models employ self-training on unlabeled datasets\cite{selfsupervised} or use vision foundation backbones such as Stable Diffusion\cite{stablediffusion} and DINOv2\cite{dinov2} to achieve scale-consistent depth predictions.
More recent works, such as DepthAnything\cite{depthanything}, leverage large-scale pretraining and temporal feature aggregation to improve continuity and accuracy across dynamic scenes.

\newcommand{\R}{\mathbb{R}}
\section{method}

\subsection{Theoretical Analysis}\label{31}
\noindent{In static environments, when the camera undergoes motion, the resulting temporal frames manifest image displacement and exposure-induced blur.}
For scene content residing on a common depth plane, the variations across different FOVs in the image adhere to a deterministic rule induced by the optical geometry.
Natural environments exhibit dense depth variation, wherein the dynamics of temporal frames are systematically linked to the depth of the respective FOV.
Under short exposure, inter-frame comparisons in the video suffice to reveal image changes, whereas for long exposures, the magnitude of change manifests as various forms of image blur.
Consequently, valuable depth information about the scene and motion parameters of the camera can be inferred from continuous video.

\begin{figure}[h]
\begin{center}
\includegraphics[width=8.3cm]{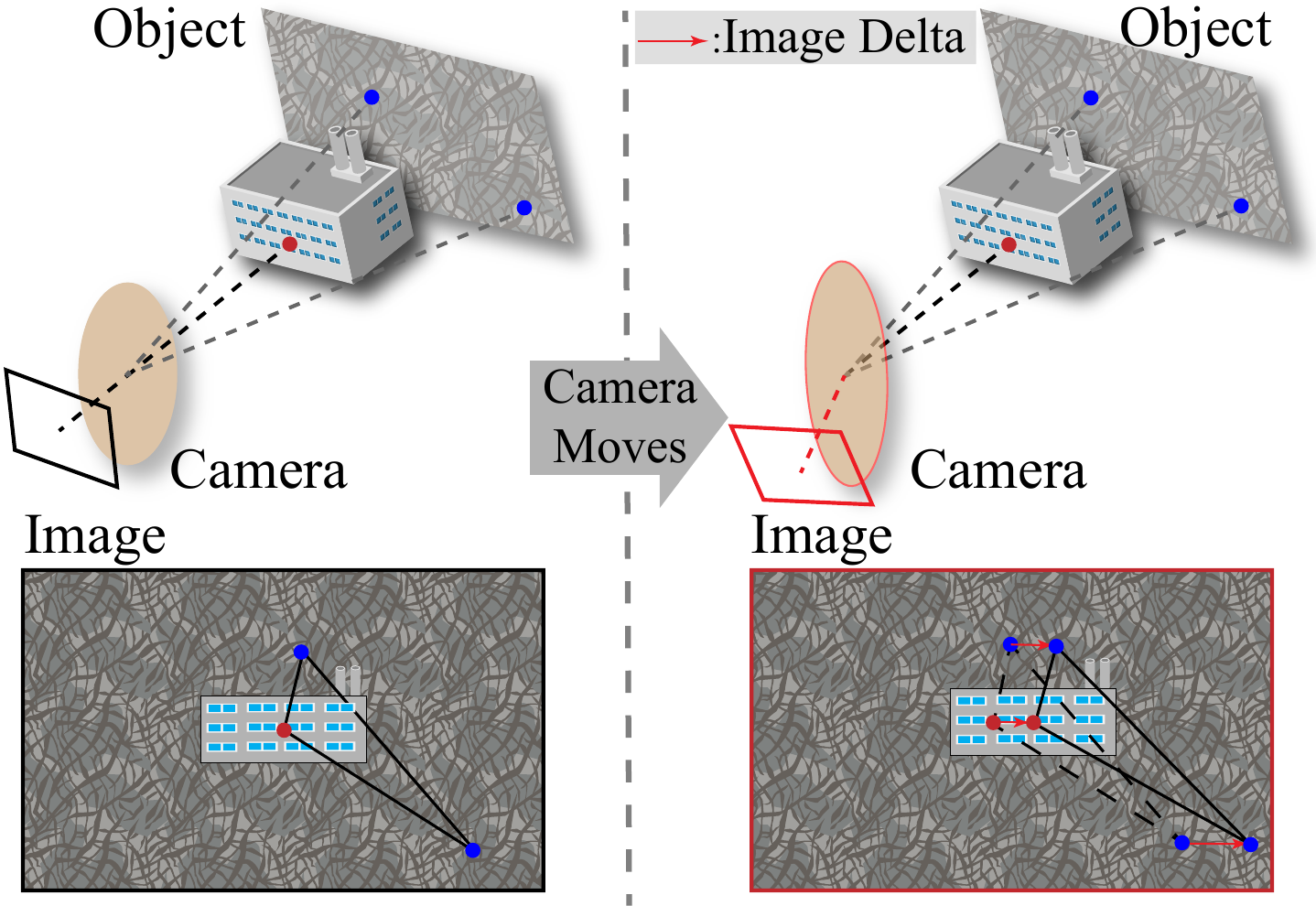}
\end{center}
\vspace{-10pt}
   \caption{An ideal optical system.  Red point is on-axis point and blue points denote off-axis point. Under an identical camera state, scene points at different spatial locations exhibit different image deltas.}
\label{fig1}
\end{figure}

Formally, given a RGB video with $T$ frames, we primarily focus on the temporal evolution of selected $N$ query points; specifically, these points are fields-of-view sampled uniformly over the first frame of the video.
As the camera shakes over time, query points shift, and their local neighborhoods exhibit particular blur patterns.
The spatio-temporal delta of query points, their corresponding object distances and camera motion trajectory are parameterized as:
\begin{itemize}
\item[$\bullet$] {\bf Delta:} $\Delta =\{\delta _i\}_{i=1}^{N}$, where each $\delta_i=\{ (p_x,p_y)_{i,t}\} _{t=1}^T$ consists of the 2D positional coordinate displacement of query point $i$ across $T$ frames.
\item[$\bullet$] {\bf Depth: $L=\{ l_i\} _{i=1}^N$}, where $l_i$ indicates the metric depth of query point $i$ at the first frame.
\item[$\bullet$] {\bf Trajectory: $\Theta=\{ \theta_t\} _{t=1}^T$}, where $\theta_t=\{ (\alpha ,\beta ,\gamma)_t\}$ encapsulates three parameters of the camera rotation at the $t$ frame.
\end{itemize}

Within an ideal optical imaging system, we consider a query point $Q_{on}$ at central FOV and another query point $Q_{off}$ at an off-axis FOV, with the sensor located at the conjugate plane of the on-axis point.
Each point is further expanded into a small neighborhood region with patch $P_{on}$ and $P_{off}$, at depth $l_{Q_{on}}$ and $l_{Q_{off}}$.
At time $t$, the camera’s rotational state is three-dimensional and we construct the plane defined by the rotation vector and the optical axis.
In this plane, the rotation angle is $\theta _t$ and the dihedral angles of this plane with respect to the image-plane x- and y-axes are $\phi _x$ and $\phi _y$, respectively.
The resulting delta onto the image plane of $Q_{on}$ is given by:
\begin{equation}\label{E31}
    \delta_{Q_{on}}=-l_{Q_{on}}tan\theta_t\frac{f'}{l_{Q_{on}}+f'},
\end{equation}
where $f'$ is the focal length of the optical system.
Meanwhile, the delta associated with $Q_{off}$ is:
\begin{equation}\label{E32}
    \delta_{Q_{off}}=-\frac{l_{Q_{on}}tan\theta_t}{l_{Q_{off}}}\frac{f'}{l_{Q_{on}}+f'}\frac{l_{Q_{off}}-y^2}{l_{Q_{off}}-ytan\theta_t},
\end{equation}
where $y$ denotes the object height corresponding to $Q_{off}$. 
Assuming small path $P_{on}$ and $P_{off}$ have locally uniform displacements $\delta_{Q_{on}}$ and $\delta_{Q_{off}}$ the imaging result of this region in $t$ frame is:
\begin{equation}\label{E33}
    G(P)=\frac{1}{t_E}\int_{0}^{t_E} \iint_{p,q\in P} I(p+(p_x)_t, q+(p_y)_t)\, dt,
\end{equation}
where $p_x = \delta_i cos\phi_x$, $p_y = \delta_i cos\phi_y$, $t_E$ is the exposure time for each frame and $I$ denotes the image under stationary condition.
It follows that the blur $\rho$ of the patches is determined by $\Delta$ and a contiguous trajectory denser than $\Theta$.
Given identical exposure settings and camera motion for both $Q_{on}$ and $Q_{off}$, the ratio of $\rho$ between them follows:
\begin{equation}\label{E34}
    \frac{\rho_{off}}{\rho_{on}}\propto\frac{\delta_{Q_{off}}}{\delta_{Q_{on}}}=-\frac{1}{l_{Q_{off}}}\frac{l_{Q_{off}}-y^2}{l_{Q_{off}}-ytan\theta_t}.
\end{equation}
From Eq. \ref{E31} and Eq. \ref{E34}, $\rho_{off}$ is associated with $\rho_{on}$, the object's spatial position and the motion trajectory, whereas $\rho_{on}$ depends only on depth and the trajectory.
Eq. \ref{E33} indicates that the blur magnitude $\rho$ depends solely on the image delta $\Delta$ under the same exposure state.

It should be emphasized that the theoretical derivation above assumes near-constant image shift over a local patch and an solid sensor pose during the exposure interval.
During depth estimation, the continuous trajectory is impractical to confirm, yet $\rho_{on}$ depends on it linearly.
In solving for dense trajectories, as temporal information is hidden during integration, blind restoration exhibits a multi-solution ambiguity.

\begin{figure}[h]
\begin{center}
\includegraphics[width=16cm]{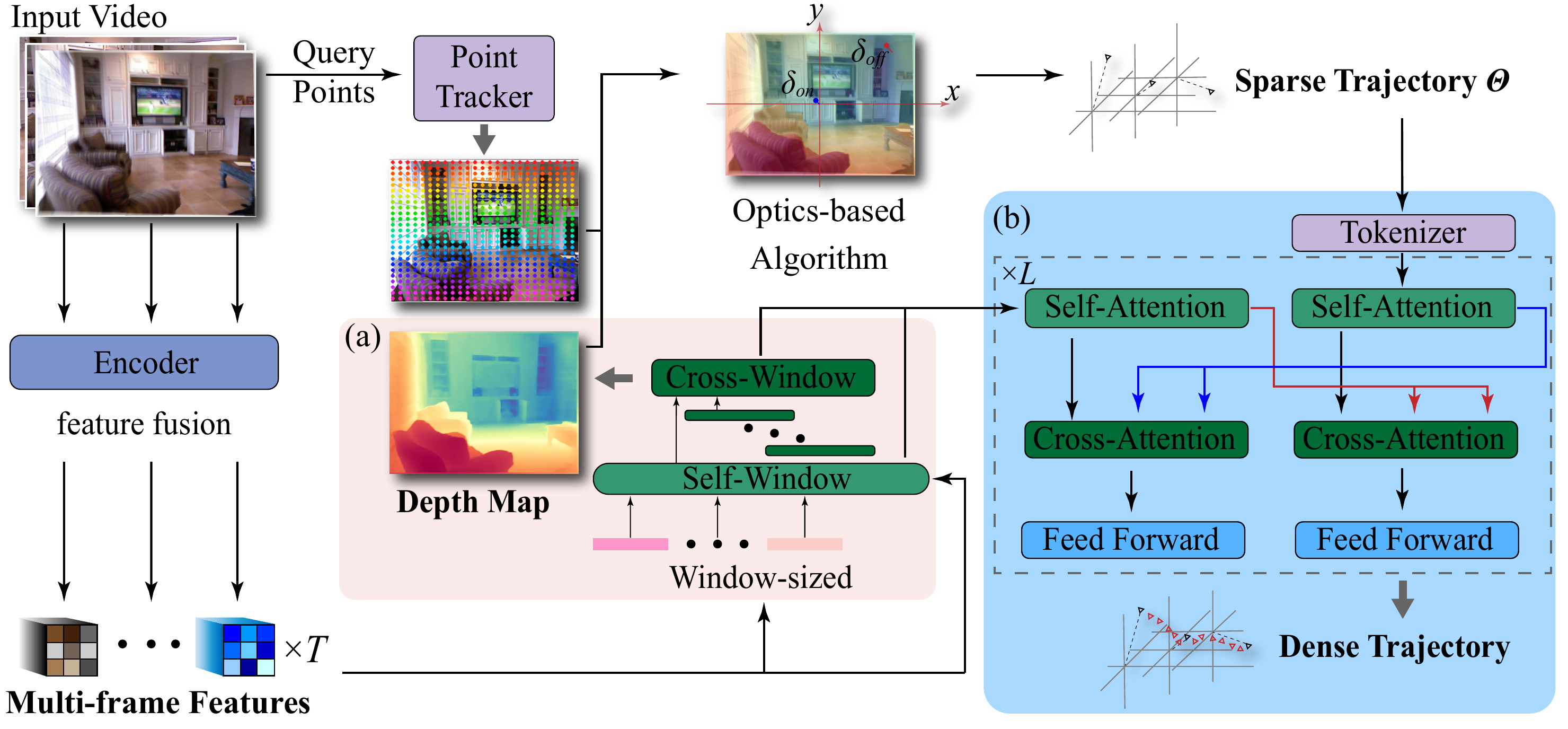}
\end{center}
\vspace{-10pt}
   \caption{{\bf Overview of our pipeline.} We begin by extracting multi-frame features with DINO\cite{dinov2} model, while employing off-the-shelf point tracker\cite{cotracker} to derive $\Delta$ of $N$ query points, from which we compute sparse trajectory $\Theta$ via optics-based algorithm. 
   {\bf (a) Depth estimation.} We segment the $T$-length features into window-size and further encode them with self-attention, followed by aggregation into the first window via cross-attention.
   {\bf (b) Dense trajectory decoder.} Multi-frame features fused with the depth map $L$ are injected into tokenized\cite{deberta} $\Theta$, resulting in the dense camera trajectory.}
\label{fig2}
\end{figure}

To address these constraints, we employ a Transformer-based framework, as shown in Figure \ref{fig2}, to extract and compare the complexities across multiple FOV regions and temporally ordered frames through learned representations.
Rather than depending exclusively on explicit equation-based computations, the approach infers metric depth of object and dense camera trajectory, thereby improving robustness to variations in motion dynamics.

\subsection{Video Information Extraction}\label{32}
\noindent{For effective depth estimation and dense trajectory computation, as discussed in Sec. \ref{31}, it's crucial that the model firstly captures the per-frame local blur $\rho$ and inter-frame motion trajectory $\Theta$.}
To address these requires, we employ DINOv2\cite{dinov2} model to obtain high-dimensional image features and off-the-shelf tracker\cite{cotracker} to estimate the temporal motion of query points.

\vspace{0.3em}\noindent{\bf Multi-frame features.}
To balance accuracy and computational complexity, we adopt DINOv2(ViT-B) as video encoder.
Larger backbones incur higher memory footprints, whereas ViT-S model with fewer attention heads diminish cross-region dependency capture, which is important for our ratio blur pattern computation. 
For enhancing scale robustness, we gather feature parameters from $K$ different layers of the encoder.
Considering a input video $V=\{I_t\}_{t=1}^T$, where $I_t\in \R^ {h\times w \times 3}$, the features extracted are $\phi_t=\{\phi _{t, k}\}_{k=1}^K$, where $\phi_{t,k}\in \R^ {h_k\times w_k \times d_k}$.
As features from different layers differ in spatial resolution and channels, low-resolution features are upsampled and multi-scale features are fused into a unified representation $\phi_t\in \R^{h_1\times w_1 \times d}$ via linear layers.
The final output multi-frame features, denoted as $\Phi=\{\phi_t\}_{t=1}^T$, aggregates per-frame features across time and scales, serving as a rich representation of our required blur pattern.

\vspace{0.3em}\noindent{\bf Query points tracking.}
In the tracking module, the pretrained point tracking model Cotracker\cite{cotracker} suffices for our single-point tracking requirement. 
For query points setting, we control its density by count of half-width, while ensuring that the on-axis point is always included.
Assuming a half-width count of $i$, the number of query points $N=(2i+1)^2$, comprising the on-axis point and other points uniformly sampled in each of the height and width directions.
The tracker model yields temporal coordinates for the query points, which we convert into delta $\Delta$ relative to the first frame.
Since the captured scene remains static and limited camera motion, the visibility term contained in the tracking model output is therefore omitted.
Subsequently, an estimation of the sparse camera trajectory $\Theta$ can be performed based on the delta $\Delta$ and the depth map $L$.

\subsection{Depth Map Estimation}\label{33}
\begin{figure}[t]
\begin{center}
\includegraphics[width=7cm]{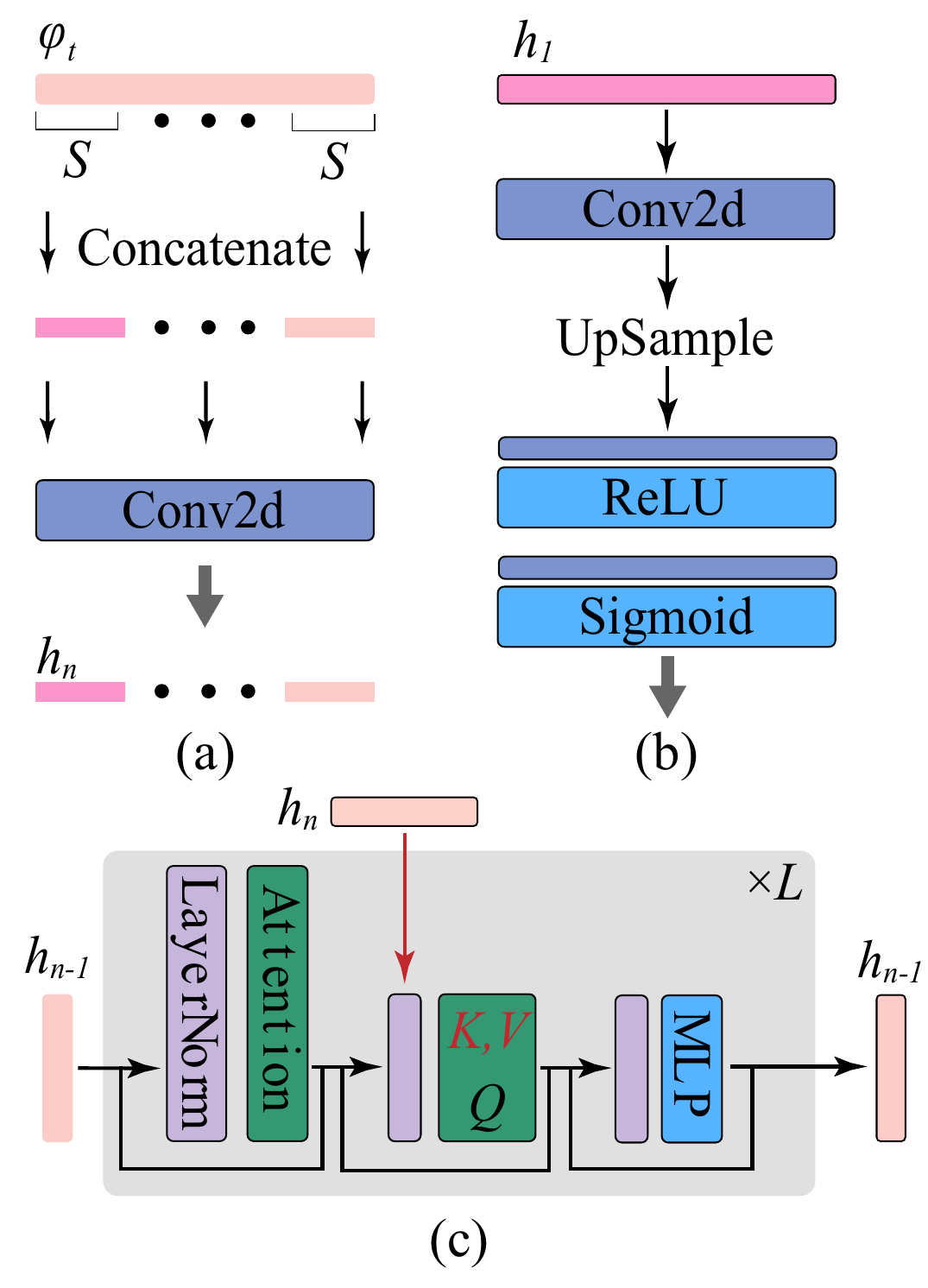}
\end{center}
\vspace{-10pt}
   \caption{{\bf (a) Window-embed.} Multi-frame features are windowed, concatenated along the channel dimension, and embeded with convolutional layer.
   {\bf (b) Output head.} The head first reduces channels by half using a convolution, then upsamples via bilinear interpolation to the original resolution, and finally employs two convolution–activation stages to produce the depth map.
   {\bf (b) $\boldsymbol Cross-Window.$} Cross-attention is computed between two windows, where the refined post-window serves as the key–value to refine the pre-window.}
\label{fig3}
\end{figure}

\noindent{In the depth estimation module, we apply attention over the extracted features and project them through the output layer to a single-channel depth map.}
The module is simplified in Figure \ref{fig2}, while the layer-wise operations are detailed in Figure \ref{fig3}.

For the input temporal features $\Phi=\{\phi_t\}_{t=1}^T$, we first apply window-embedding to further aggregate ambiguous information.
With short exposure time recording, omitting window-embedding leads to performance degradation in the downstream attention computation.
Specifically, let $S$ denotes the window length, $\{\phi_t\}_{t=(n-1)S}^{nS}$ are concatenated along the channel dimension, and then use convolutional layers to reduce the channels back to the original size $\phi_t\in \R^{h_1\times w_1 \times d}$.

Following window-embedding, the features are processed by the $Self-Window$ module, which is invoked again during trajectory densification, and the parameters are shared across both uses.
This module primarily compares different local regions within the same frame to extract relative depth.
In $Self-Window$, we use stacked self-attention and feed-forward layers as a decoder without cross-attention to the encoder, i.e., a decoder-only design.

The $Cross-Window$ module aggregates inter-frame information and infers metric depth by tracking changes of the same region across recording time.
Temporally, we apply the module to windows in a backward incremental manner, proceeding from later to earlier frames until the first window.
Theoretically, the mapping between blur change and depth is frame-count invariant; consequently, the window count governs invocation frequency but leaves the module’s weights solid.

Formally, the multi-frame features are embedded and processed through $Self-Window$ and $Cross-Window$ modules:
\begin{align}\label{E35}
    & h^0_n = Embedding(\{\phi_t\}_{t=(n-1)S}^{nS}), n=1,...,T//S, 
    \\& h_n^l=Self-Window^l(h^{l-1}_n), l=1,...,L,
    \\& h_{n-1}^C=Cross-Window(h_n^C, h_{n-1}^L), n=T//S,...,2,
\end{align}
where $h_1^L$ is the final output after the attention computation and $h_{T//S-1}^C=Cross-Window(h_{T//S}^L,h^L_{T//S-1})$.
The final output head, composed of convolution layers and upsampling, compresses features to one channel and recovers the native spatial resolution to output the depth map $L$.

\subsection{Trajectory Generation Algorithm.}\label{34}
\noindent{Camera trajectory estimation proceeds in two stages: geometry-based inter-frame motion computation followed by attention-based dense trajectory generation.}
For the former stage, we take the tracker outputs $\Delta$ and the depth map $L$ as inputs; for the latter, we further reuse the multi-frame features additionally.

\vspace{0.3em}\noindent{{\bf Optics-based algorithm.}}
The trajectory $\theta_t$ includes three angles (e.g., roll, pitch, yaw), corresponding to the camera’s three rotational DOFs. 
The algorithm computes roll, yaw, and pitch sequentially, since the latter angles are conditioned on the former.

We first determine the roll from the observed image rotation, since the induced delta is antisymmetric across the FOVs.
Concretely, we compute the mean vertical offset $p_y$ over all query points along $y=0$ FOV.
As the y-displacement on this FOV is invariant to the x-field, this statistic yields the displacement with the roll component removed.
Subtracting this mean from the previously offsets, we then combine the residuals with their coordinates to recover the roll angle.

Subsequently, we consider the $p_x$ of $x=0$ FoV line.
By subtracting, for each position, the displacement induced by the recovered roll motion, the remaining delta yields the yaw angle.
Since the superposition of yaw and pitch contaminates $p_y$, only $p_x$ remains discriminative for their estimation.

In the final step, we recover the pitch predominantly from $p_y$ of all query points, leveraging the FoV coordinates in both x and y, together with the previously estimated yaw degree.
The final output is formed via a weighted average across FOVs and weights are linearly tapered with field extent due to the aberration level increasing toward the periphery.

Computational efficiency is improved via variable vectorization, i.e., rewriting the pipeline in matrix form to minimize per-sample loops.
As the angles are defined post long-term temporal coupling, we resort to an optical computational method that sidesteps explicit disentanglement.
Neural-network-based decoupling would be constrained by the practical difficulty of assembling a suitable dataset.

\vspace{0.3em}\noindent{{\bf Dense trajectory decoder.}}
For the dense trajectory estimation, we employ a text–image fusion paradigm in which linguistic information is integrated with visual representations to obtain trajectory fields.

Sparse trajectory are up-sampled using linear interpolation to ensure temporal consistency, after which the sequences are tokenized\cite{deberta} for downstream modeling.
Instead of window-embedding, we process the image features through the self-window block directly, after which we concatenate them with the final decoder output of the depth estimation part before output head.
With both image and trajectory-text embeddings in hand, we leverage a vision–language framework to integrate the two, resulting in higher-precision dense trajectory fields.

While dense trajectory can, in theory, be inferred solely from multi-frame features, our experiments indicate that incorporating depth features (via concatenation) both speeds convergence and enhances reconstruction accuracy.
Additionally, after upsampling of $\Theta$, we ensure that the text sequence length matches the output sequence length, and no sampling is performed within the feed-forward layer.
\section{Experiment}

\begin{figure*}
\begin{center}
\includegraphics[width=\linewidth]{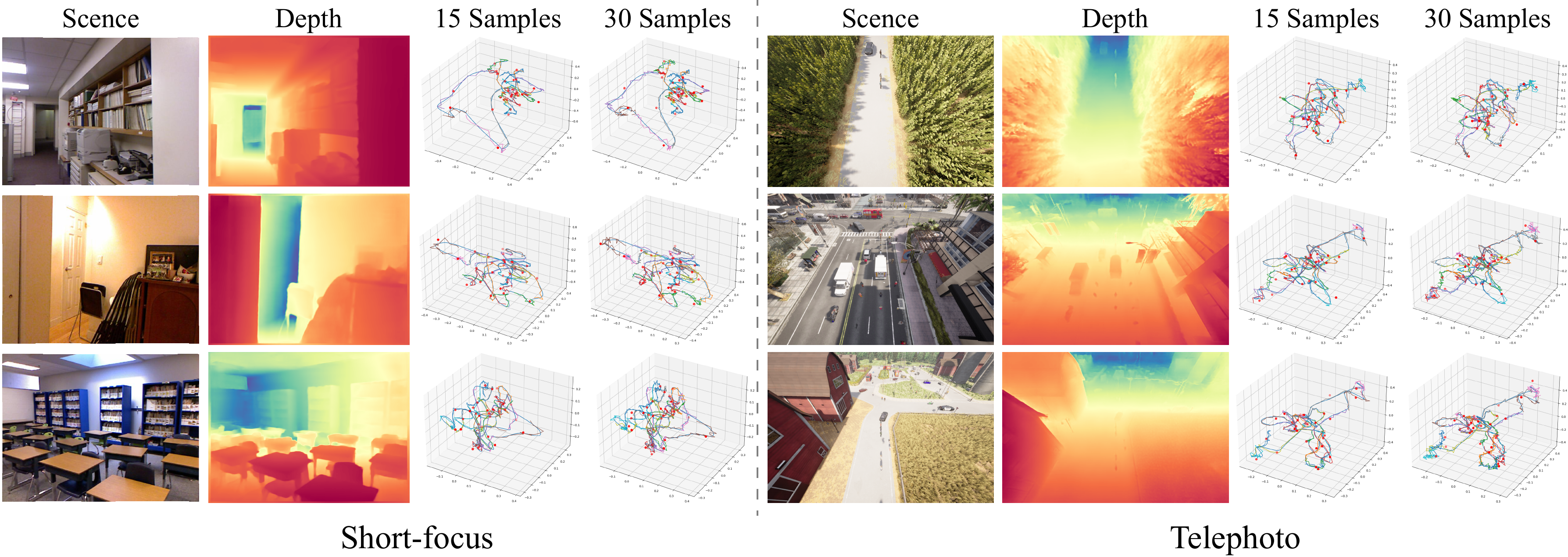}
\end{center}
\vspace{-10pt}
   \caption{Results of our pipeline, including depth estimation and trajectory reconstruction. Here, samples denotes the number of samples within a single frame of the reconstructed dense trajectory. In the 3D plot, we render the GT trajectory as a blue solid line, depict the sparse trajectory with red markers, and use distinct colors to represent the reconstructed trajectory for each individual frame.}
\label{fig4}
\end{figure*}

\subsection{Datasets and Evaluation Protocol}\label{41}
\noindent{{\bf Datasets.}} Our scenario-oriented video data are derived from depth datasets through a imaging simulation pipeline based on Sec. \ref{31}.
In order to cover a broad focal range and to model the distinct optical formulations of periscope telephoto systems, we employ indoor dataset NYU Depth\cite{nyudepth} for short-focus camera and synthetic outdoor dataset SKYScenes\cite{SKYSCENES} exhibiting a larger depth extent for telephoto one.
Furthermore, leveraging a densely sampled camera motion and RGB–depth pairs, we produce videos with configurable exposure times and frame budgets, detailed in our supplementary material.

The camera trajectory is sourced from a gyroscope-recorded, long-duration handheld sequence with a 2-ms sampling interval as shown at the bottom left of Figure \ref{fig0}.
Our trajectory sequence is 67.14 seconds long; we randomize the exposure onset for each video, constrained to preserve the prescribed exposure time.
The video is rendered with a 60 ms per-row exposure, a 4 µs per-row transfer time, and 30 frames overall.
For NYU Depth dataset, we select 1,449 densely annotated images, with 200 scenes held out for validation.
And in the SkyScences dataset, we select ClearNoon scenes with heights between 15 m and 35 m and pitch angles from 0° to 60°.
Post filtering, a set of 2,151 object-rich images is curated, from which 200 images are randomly selected to form the validation subset.
The resulting corpus includes 3,200 training videos and 400 validation videos, both with its start-time of exposure.
Samples are annotated as short-focus or telephoto camera, acknowledging that the optics-based algorithms in Sec. \ref{34}  differ.

\vspace{0.3em}\noindent{{\bf Evaluation protocol.}} Within the pipeline, we perform separate training and validation for the depth estimation module in Sec. \ref{33} and the trajectory estimation module in Sec. \ref{34}.
Depth estimation is validated using established metrics and comparative baselines.
In contrast, we measure the accuracy of the camera trajectory using following metrics.

We align the estimated dense camera trajectory to the simulated target one and report the L1 error over the trajectory samples.
{\bf AbsRel} is used to quantify proportional trajectory error, the same as depth part, L1 error over the target trajectory.
To assess the accuracy of trajectory predictions under different thresholds, we utilize {\bf Accuary} $e_i, i=1, 2$, and unlike depth evaluation, the criterion mainly involves {\bf AbsRel} instead of the ratio of the predicted trajectory to the target one.

\subsection{Implementation Details}
\noindent{We decouple the training pipeline by optimizing depth estimation and trajectory estimation in two separate stages.}
For the former, we employ full-parameter training except the point tracker, whereas for the latter, all weights are frozen except those within the in Figure \ref{fig2} (b).
We train both stage using the AdamW optimizer\cite{adamw} with a learning rate of $1\times 10^{-3}$ and a weight decay of $5\times 10^{-5}$.
The number of iterations in the two stages of training is 320$K$, with warm-up step of 1000.
We employ one NVIDIA RTX 4090 GPU for training, with a batch size of 8.
We use the number of Transformer layers, where layer has a hidden dimension of 384 and attention heads of 8, for $Self-Window$, $Cross-Window$ and the trajectory decoder in Sec. \ref{34} as $L=6, 2,6$, respectively.
In dataset preparation, we standardize the input resolution to 518 $\times$ 518; for depth estimation, the tempporal window parameter is configured as $S=6$.
For point tracker, we set the query points grid as 25 $\times$ 25, $N=625$ and conduct inference under the offline mode.

\subsection{Main Results}

\noindent{{\bf Depth map results.}} We conduct a systematic evaluation of the proposed depth estimation method on three benchmarks; the quantitative metrics and comparative results are presented in Table \ref{table:depth}. 
Two validation datasets are derived from our internally constructed validation set in Sec. \ref{41}.
For the KITTI\cite{kitti} benchmark, we identify 10 scenes featuring comparatively dense depth annotations and select several single high-clarity frames from each scene to synthesize videos, resulting in a total of 300 samples.
For comparison, we apply single-image depth estimators\cite{depthanything, zoedepth, IEBins, P3Depth} to the first video frame, and, for video depth estimators\cite{DepthCrafter, chronodepth}, we extract the first frame from their predictions, ensuring frame-aligned evaluation.

\begin{table*}[h]
    \centering
    \caption{\textbf{Quantitative results of depth map estimation.} The best one is emphasized using bold type, and the second one is indicated via underlined text.}
    \vspace{-3pt}
    \resizebox{\textwidth}{!}{
    \setlength{\tabcolsep}{4pt}
    \begin{tabular}{l|ccc|ccc|ccc|ccc}
    \toprule
    Depth & \multicolumn{3}{c|}{NYU-D~\cite{nyudepth}}
          & \multicolumn{3}{c|}{SKYScenes~\cite{SKYSCENES}}
          & \multicolumn{3}{c|}{KITTI~\cite{kitti}}
          & \multicolumn{3}{c}{\textbf{Average}} \\
    Estimator
          & AbsRel $\downarrow$ & $\delta_1$  $\uparrow$ & $\delta_2$  $\uparrow$
          & AbsRel $\downarrow$ & $\delta_1$  $\uparrow$ & $\delta_2$  $\uparrow$
          & AbsRel $\downarrow$ & $\delta_1$  $\uparrow$ & $\delta_2$  $\uparrow$
          & AbsRel $\downarrow$ & $\delta_1$  $\uparrow$ & $\delta_2$  $\uparrow$ \\
    \midrule\midrule

    \multicolumn{13}{c}{\textit{Single Image Depth Estimator}}\\
    
    P3Depth~\cite{P3Depth}
      & 0.104  & 0.898 & 0.981
      & 0.155 & 0.817 & 0.898
      & 0.071  & 0.953 & 0.993
      & 0.110  & 0.889 & 0.957 \\
    \midrule

    IEBins~\cite{IEBins}
      & 0.087  & 0.936 & 0.992
      & 0.130 & 0.849 & 0.912
      & \underline{0.050}  & \underline{0.978} & \textbf{0.998}
      & 0.089  & 0.921 & 0.967 \\
    \midrule

    ZoeDepth~\cite{zoedepth}
      & 0.077  & 0.951 & 0.994
      & 0.115 & 0.856 & \underline{0.950}
      & 0.054  & 0.971 & 0.996
      & 0.082  & 0.926 & \underline{0.980}   \\
    \midrule

    Depth-Anything~
      & \multirow{2}{*}{\underline{0.063}} & \multirow{2}{*}{\underline{0.977}} & \multirow{2}{*}{\underline{0.997}}
      & \multirow{2}{*}{0.104} & \multirow{2}{*}{\textbf{0.888}} & \multirow{2}{*}{0.927}
      & \multirow{2}{*}{\textbf{0.048}} & \multirow{2}{*}{\textbf{0.979}} & \multirow{2}{*}{\textbf{0.998}}
      & \multirow{2}{*}{\underline{0.072}} & \multirow{2}{*}{\underline{0.948}} & \multirow{2}{*}{0.974} \\
    (ViT-B)~\cite{depthanything}
      & & &
      & & &
      & & & \\
    \midrule
    \midrule

    \multicolumn{13}{c}{\textit{Video Depth Estimator}}\\

    DepthCrafter~\cite{DepthCrafter}
      & 0.072  & 0.948 & 0.994
      & 0.106 & 0.838 & 0.921
      & 0.064  & 0.913 & 0.984
      & 0.081  & 0.900 & 0.966 \\
    \midrule
    
    ChronoDepth~\cite{chronodepth}
      & 0.070  & 0.936 & 0.992
      & \underline{0.100} & 0.871 & 0.948
      & 0.073  & 0.956 & 0.992
      & 0.081  & 0.921 & 0.977 \\
    \midrule
    
    \rowcolor{gray!20} (Ours)
      & \textbf{0.054} & \textbf{0.989} & \textbf{0.998}
      & \textbf{0.095} & \underline{0.884} & \textbf{0.953}
      & 0.051 & \underline{0.978} & \textbf{0.998}
      & \textbf{0.067} & \textbf{0.950} & \textbf{0.983} \\

    \bottomrule
    \end{tabular}}
    \label{table:depth}
    \vspace{-5pt}
\end{table*}

Table \ref{table:depth} presents a comprehensive comparative study against competing approaches.
For fairness, Depth-Anything is evaluated using its ViT-B model, which is aligned with the capacity of our DINO encoder.
Our method outperforms other baselines in the NYU Depth dataset, while achieves strong results on SKYScenes and KITTI.
The evidence suggests that the proposed method scales effectively to heterogeneous scenes and maintains competitive performance, underscoring its superior generalization and robustness.
Nevertheless, in our prior expectation, synthetic datasets, which characterized by larger depth ranges and sharper depth discontinuities, would exhibit more pronounced blur patterns, thereby yielding superior estimation performance.
We conjecture that the inverse experimental result is attributable to the limited FOV of telephoto optics, as the absence of wide-angle context reduces the available reference information.

\begin{table*}[h]
    \centering
    \caption{\textbf{Quantitative results of trajectory in different scale.} $\Theta$ is contrasted with canonical SfM methods, while the dense trajectory is validated against ground truth at different sampling densities. \textcolor{gray}{Post-interpolation} refers to $\Theta$ after upsampling; the results indicate that our model yields a tangible gain in trajectory accuracy.}
    \vspace{-3pt}
    \resizebox{\textwidth}{!}{%
    \setlength{\tabcolsep}{4pt}
    \begin{tabular}{l|ccc|ccc|ccc|ccc}
    \toprule
    \multirow{2}{*}{Methods} & \multicolumn{3}{c|}{NYU-D~\cite{nyudepth}}
          & \multicolumn{3}{c|}{SKYScenes~\cite{SKYSCENES}}
          & \multicolumn{3}{c|}{KITTI~\cite{kitti}}
          & \multicolumn{3}{c}{\textbf{Average}} \\

          & AbsRel $\downarrow$ & $e_1$  $\uparrow$ & $e_2$  $\uparrow$
          & AbsRel $\downarrow$ & $e_1$  $\uparrow$ & $e_2$  $\uparrow$
          & AbsRel $\downarrow$ & $e_1$  $\uparrow$ & $e_2$  $\uparrow$
          & AbsRel $\downarrow$ & $e_1$  $\uparrow$ & $e_2$  $\uparrow$ \\
    \midrule\midrule

    \multicolumn{13}{c}{\textit{Inter-frame Trajectory $\Theta$}}\\
    
    COLMAP~\cite{colmap}
      & 0.486  & 0.228 & 0.262
      & 0.581 & 0.212 & 0.270
      & 0.489  & 0.263 & 0.338
      & 0.519  & 0.234 & 0.290 \\
    \midrule

    DUSt3R~\cite{dust3r_cvpr24}
      & 0.211  & 0.633 & 0.865
      & 0.215 & 0.590 & 0.777
      & 0.210  & 0.610 & 0.792
      & 0.212  & 0.611 & 0.811 \\
    \midrule

    DiffusionSfM~\cite{diffusionsfm}
      & \underline{0.209}  & \underline{0.643} & \underline{0.877}
      & \underline{0.161}  & \underline{0.755} & \underline{0.934}
      & \underline{0.164}  & \underline{0.748} & \underline{0.919}
      & \underline{0.178}  & \underline{0.715} & \underline{0.910}   \\
    \midrule

    \rowcolor{gray!20} Ours + CoTracker\cite{cotracker}
      & \textbf{0.045} & \textbf{0.933} & \textbf{1.000}
      & \textbf{0.043} & \textbf{0.978} & \textbf{1.000}
      & \textbf{0.045} & \textbf{0.967} & \textbf{1.000}
      & \textbf{0.044} & \textbf{0.959} & \textbf{1.000}   \\
    \midrule
    \midrule

    \multicolumn{13}{c}{\textit{Dense Trajectory}}\\
    \multicolumn{13}{c}{\textit{15 per Frame}}\\

    \textcolor{gray}{Post-interpolation}
      & \textcolor{gray}{0.285} & \textcolor{gray}{0.284} & \textcolor{gray}{0.745}
      & \textcolor{gray}{0.329} & \textcolor{gray}{0.373} & \textcolor{gray}{0.754}
      & \textcolor{gray}{0.305} & \textcolor{gray}{0.292} & \textcolor{gray}{0.688}
      & \textcolor{gray}{0.306} & \textcolor{gray}{0.316} & \textcolor{gray}{0.729} \\
    \midrule
    
    Ours + LocoTracker\cite{locotracker}
      & 0.068  & 0.803 & 0.880
      & \underline{0.069}  & 0.818 & 0.907
      & 0.077  & 0.745 & 0.836
      & 0.071  & 0.789 & 0.874 \\
    \midrule
    
    \rowcolor{gray!20} Ours + CoTracker\cite{cotracker}
      & \underline{0.064} & \underline{0.865} & \underline{0.949}
      & 0.070 & \underline{0.858} & \underline{0.951}
      & \underline{0.070} & \underline{0.851} & \underline{0.955}
      & \underline{0.068} & \underline{0.858} & \underline{0.952} \\
    \midrule

    \multicolumn{13}{c}{\textit{30 per Frame}}\\
    
    \textcolor{gray}{Post-interpolation}
      & \textcolor{gray}{0.315} & \textcolor{gray}{0.353} & \textcolor{gray}{0.719}
      & \textcolor{gray}{0.402} & \textcolor{gray}{0.196} & \textcolor{gray}{0.575}
      & \textcolor{gray}{0.313} & \textcolor{gray}{0.218} & \textcolor{gray}{0.657}
      & \textcolor{gray}{0.343} & \textcolor{gray}{0.256} & \textcolor{gray}{0.650} \\
    \midrule
    
    Ours + LocoTracker\cite{locotracker}
      & 0.066  & 0.825 & 0.908
      & \textbf{0.067}  & 0.841 & 0.925
      & 0.071  & 0.790 & 0.878
      & \underline{0.068}  & 0.819 & 0.904 \\
    \midrule
    
    \rowcolor{gray!20} Ours + CoTracker\cite{cotracker}
      & \textbf{0.063} & \textbf{0.867} & \textbf{0.955}
      & \underline{0.069} & \textbf{0.866} & \textbf{0.952}
      & \textbf{0.068} & \textbf{0.861} & \textbf{0.957}
      & \textbf{0.067} & \textbf{0.865} & \textbf{0.955} \\

    \bottomrule
    \end{tabular}
    }
    \label{table:trajectory}
    \vspace{-5pt}
\end{table*}

\vspace{0.3em}\noindent{{\bf Trajectory results.}} The trajectory evaluation follows the same datasets as described above, with the ground truth defined by the trajectory curves in Figure \ref{fig4}.
Due to scale discrepancies in the estimated results, we average the angle values within one frame for inter-frame trajectory; and for the 15 samples groups, we subsample the ground truth.

Table \ref{table:trajectory} presents the comparisons against SfM methods, delineates the differences among trackers, and quantifies the performance gains contributed by our dense trajectory decoder. 
Although post-averaging inter-frame trajectory may introduce deviations from the true values, our method substantially outperforms competing approaches.
Regarding the tracker, CoTracker\cite{cotracker} exhibits a measurable advantage over LocoTracker\cite{locotracker}; accordingly, we adopt the former in our pipeline.

\subsection{Ablation Studies}

\noindent{We present our ablation studies on Table \ref{tab:ablation}.}
We first ablate the design effectiveness of the depth estimation stage.
In \textbf{(I)}, the window size is systematically adjusted, encompassing reduced-length and extended-length settings.
As changing $S$ alters the number of concatenated features, we accordingly adjust the convolutional layer configuration during training.
In \textbf{(II)}, we ablate the $Cross-Window$ mechanism by removing it and directly feeding the output of $Self-Window$, into the output head.
In \textbf{(III)}, we evaluate varying number of Transformer layers in our dense trajectory decoder.

The observed performance degradation demonstrates the effectiveness of our pipeline design.
Specifically, in \textbf{(I)}, shorter $S$ yield superior performance compared to longer one.
In \textbf{(II)}, removing the $Cross-Window$ module substantially degrades depth accuracy and, in turn, diminishes the quality of the trajectory reconstruction.
Finally, \textbf{(III)} shows that: within 6 layers, reducing the depth of the network consistently degrades performance; therefore, balancing accuracy with model size, we set the number of layers as 6.

\begin{table}[h]
    \centering
    \caption{\textbf{Ablation studies.} We evaluate performance of different baselines on both depth estimation and trajectory metrics.}
    \vspace{0pt}
    \resizebox{10cm}{!}{
    \setlength{\tabcolsep}{3pt}
    \begin{tabular}{l|cc|c|ccc|c}
    \toprule
    Average 
    & \multicolumn{2}{c|}{\textbf{(I) $S$} } 
    & \textbf{(II)}
    & \multicolumn{3}{c|}{\textbf{(III)} $L$} 
    & \multirow{2}{*}{Ours} \\

    Metrics
    & 4 & 8
    & $\bcancel{cross-}$
    & 3 & 4 & 5
    &  \\
    \midrule
    \midrule

    $\delta_1 \uparrow$
    & \underline{0.946} & 0.923
    & 0.887
    & - & -  & - 
    & \textbf{0.950}  \\
    \midrule
    \midrule
    
    AbsRel $\downarrow$
    & \underline{0.071} & 0.078
    & 0.091
    & 0.082 & 0.077 & 0.073
    & \textbf{0.067}  \\
    \midrule

    $e_1 \uparrow$ 
    & 0.862 & 0.812
    & 0.810
    & 0.783 & 0.845 & \underline{0.863}
    & \textbf{0.865}  \\
    \bottomrule
    \end{tabular}}
    \label{tab:ablation}
    \vspace{-10pt}
\end{table}

\section{Conclusion}
\label{sec:conclusion}

We introduce a novel perspective-blur-based framework for depth estimation and trajectory densification that utilizing blur cues and temporal image variations present in video sequences.
Leveraging attention-based encoder and decoders, we reconstruct single, dense metric depth map and infer the corresponding camera trajectory directly from monocular video input.
This approach offers substantial utility in scenarios necessitating dense camera pose information, for instance, in assessing camera stabilization.
We aspire for this work to inspire the community to pursue broader innovation and exploration in related fields.

\bibliographystyle{unsrt}  
\bibliography{references}

@String(CVPR= {IEEE Conf. Comput. Vis. Pattern Recog.})

@String(ICCV= {Int. Conf. Comput. Vis.})

@String(ECCV= {Eur. Conf. Comput. Vis.})

@String(NIPS= {Adv. Neural Inform. Process. Syst.})

@String(ICLR = {Int. Conf. Learn. Represent.})

@String(IJCAI = {IJCAI})

@String(CVPR  = {CVPR})

@String(ICCV  = {ICCV})

@String(ECCV  = {ECCV})

@String(NIPS  = {NeurIPS})

@String(ICLR  = {ICLR})

@article{longuet-higgins_computer_1981,
	title = {A computer algorithm for reconstructing a scene from two projections},
	volume = {293},
	number = {5828},
	journal = {Nature},
	author = {Longuet-Higgins, H. C.},
	month = sep,
	year = {1981},
	pages = {133--135},
}

@article{3DCV,
author = {Faugeras, Olivier},
year = {1993},
month = {01},
pages = {},
title = {Three-dimensional computer vision: a geometric viewpoint},
journal = {MIT press}
}

@article{Multiple_View_Geometry,
author = {Page, G.},
year = {2005},
month = {03},
pages = {271},
title = {Multiple View Geometry in Computer Vision, by Richard Hartley and Andrew Zisserman, CUP, Cambridge, UK, 2003, vi+560 pp., ISBN 0-521-54051-8.},
volume = {23},
journal = {Robotica},
}

@article{BA,
author = {Triggs, B. and Mclauchlan, Philip and Hartley, R. and Fitzgibbon, Andrew},
year = {2000},
month = {01},
pages = {198-372},
title = {Bundle adjustment - A modern synthesis},
journal = {ICCV '99 Proceedings of the International Workshop on Vision Algorithms: Theory and Practice}
}

@article{photo_tourism,
author = {Snavely, Noah and Seitz, Steven and Szeliski, Richard},
year = {2006},
month = {01},
pages = {835-846},
title = {Photo tourism: Exploring photo collections in 3D},
volume = {25},
journal = {ACM Transactions on Graphics (2006)}
}

@inproceedings{visual_sfm,
author = {Wu, Changchang},
year = {2013},
month = {06},
pages = {127-134},
title = {Towards Linear-Time Incremental Structure from Motion},
journal = {Proceedings - 2013 International Conference on 3D Vision, 3DV 2013},
}

@inproceedings{colmap,
author = {Schönberger, Johannes and Frahm, Jan-Michael},
year = {2016},
month = {06},
pages = {},
title = {Structure-from-Motion Revisited},
journal = {CVPR},
}

@article{Fischler1981RandomSC,
  title={Random sample consensus: a paradigm for model fitting with applications to image analysis and automated cartography},
  author={Martin A. Fischler and Robert C. Bolles},
  journal={Commun. ACM},
  year={1981},
  volume={24},
  pages={381-395},
}

@article{8point,
  author={Hartley, R.I.},
  journal={IEEE Transactions on Pattern Analysis and Machine Intelligence}, 
  title={In defense of the eight-point algorithm}, 
  year={1997},
  volume={19},
  number={6},
  pages={580-593},
}

@article{SBA,
author = {Lourakis, Manolis and Argyros, Antonis},
year = {2009},
month = {01},
pages = {},
title = {SBA: A Software Package for Generic Sparse Bundle Adjustment},
volume = {36},
journal = {ACM Trans. Math. Softw.}
}

@INPROCEEDINGS{1Dsfm,
  author={Crandall, David and Owens, Andrew and Snavely, Noah and Huttenlocher, Dan},
  booktitle={CVPR 2011}, 
  title={Discrete-continuous optimization for large-scale structure from motion}, 
  year={2011},
  volume={},
  number={},
  pages={3001-3008},
}

@inproceedings{theiasfm,
author = {Sweeney, Chris and Höllerer, Tobias and Turk, Matthew},
year = {2015},
month = {10},
pages = {693-696},
title = {Theia: A Fast and Scalable Structure-from-Motion Library},
}

@inproceedings{openMVG,
author = {Moulon, Pierre and Monasse, Pascal and Perrot, Romuald and Marlet, Renaud},
year = {2017},
month = {04},
pages = {60-74},
title = {OpenMVG: Open Multiple View Geometry},
isbn = {978-3-319-56413-5},
}

@inbook{deepsfm,
author = {Wei, Xingkui and Zhang, Yinda and Li, Zhuwen and Fu, Yanwei and Xue, Xiangyang},
year = {2020},
month = {11},
pages = {230-247},
title = {DeepSFM: Structure from Motion via Deep Bundle Adjustment},
isbn = {978-3-030-58451-1},
}

@article{Tang2018BANet,
  title={BA-Net: Dense Bundle Adjustment Network},
  author={Chengzhou Tang and Ping Tan},
  journal={ArXiv},
  year={2018},
  volume={abs/1806.04807},
}

@inproceedings{lin2021barf,
  title={BARF: Bundle-Adjusting Neural Radiance Fields},
  author={Lin, Chen-Hsuan and Ma, Wei-Chiu and Torralba, Antonio and Lucey, Simon},
  booktitle={IEEE International Conference on Computer Vision ({ICCV})},
  year={2021}
}

@inproceedings{vggsfm,
author = {Wang, Jianyuan and Karaev, Nikita and Rupprecht, Christian and Novotny, David},
year = {2024},
month = {06},
pages = {21686-21697},
title = {VGGSfM: Visual Geometry Grounded Deep Structure from Motion},
}

@misc{mast3r_eccv24,
      title={Grounding Image Matching in 3D with MASt3R}, 
      author={Vincent Leroy and Yohann Cabon and Jerome Revaud},
      booktitle = {ECCV},
      year = {2024}
}

@inproceedings{dust3r_cvpr24,
      title={DUSt3R: Geometric 3D Vision Made Easy}, 
      author={Shuzhe Wang and Vincent Leroy and Yohann Cabon and Boris Chidlovskii and Jerome Revaud},
      booktitle = {CVPR},
      year = {2024}
}

@inproceedings{ijcai,
author = {Lucas, Bruce and Kanade, Takeo},
year = {1981},
month = {04},
pages = {},
title = {An Iterative Image Registration Technique with an Application to Stereo Vision (IJCAI)},
volume = {81},
journal = {[No source information available]}
}

@article{goodfeature,
author = {Shi, Jianbo and Tomasi, Carlo},
year = {2000},
month = {03},
pages = {},
title = {Good Features to Track},
volume = {600},
journal = {Proceedings / CVPR, IEEE Computer Society Conference on Computer Vision and Pattern Recognition. IEEE Computer Society Conference on Computer Vision and Pattern Recognition},
}

@article{active,
author = {Matthews, Iain and Baker, Simon},
year = {2004},
month = {03},
pages = {},
title = {Active Appearance Models Revisited},
volume = {60},
journal = {International Journal of Computer Vision},
}

@article{opticalflow,
author = {Horn, Berthold and Schunck, Brian},
year = {1981},
month = {08},
pages = {185-203},
title = {Determining Optical Flow},
volume = {17},
journal = {Artificial Intelligence},
}

@inproceedings{KLTtracker,
  title={Pyramidal implementation of the lucas kanade feature tracker},
  author={J.-Y. Bouguet},
  year={1999},
}

@article{sift,
  title={Distinctive Image Features from Scale-Invariant Keypoints},
  author={David G. Lowe},
  journal={International Journal of Computer Vision},
  year={2004},
  volume={60},
  pages={91-110},
}

@inproceedings{orb,
author = {Rublee, Ethan and Rabaud, Vincent and Konolige, Kurt and Bradski, Gary},
year = {2011},
month = {11},
pages = {2564-2571},
title = {ORB: an efficient alternative to SIFT or SURF},
journal = {Proceedings of the IEEE International Conference on Computer Vision},
}

@inproceedings{superpoint,
author = {DeTone, Daniel and Malisiewicz, Tomasz and Rabinovich, Andrew},
year = {2018},
month = {06},
pages = {337-33712},
title = {SuperPoint: Self-Supervised Interest Point Detection and Description},
}

@article{locotracker,
  title={Local All-Pair Correspondence for Point Tracking},
  author={Cho, Seokju and Huang, Jiahui and Nam, Jisu and An, Honggyu and Kim, Seungryong and Lee, Joon-Young},
  journal={arXiv preprint arXiv:2407.15420},
  year={2024}
}

@article{tapnet,
  title={TAP-Vid: A Benchmark for Tracking Any Point in a Video},
  author={Carl Doersch and Ankush Gupta and Larisa Markeeva and Adri{\`a} Recasens and Lucas Smaira and Yusuf Aytar and Jo{\~a}o Carreira and Andrew Zisserman and Yezhou Yang},
  journal={ArXiv},
  year={2022},
  volume={abs/2211.03726},
}

@INPROCEEDINGS{pip,
  author={Nauta, Meike and Schlötterer, Jörg and van Keulen, Maurice and Seifert, Christin},
  booktitle={2023 IEEE/CVF Conference on Computer Vision and Pattern Recognition (CVPR)}, 
  title={PIP-Net: Patch-Based Intuitive Prototypes for Interpretable Image Classification}, 
  year={2023},
  volume={},
  number={},
  pages={2744-2753},
}

@article{tapir,
  title={TAPIR: Tracking Any Point with per-frame Initialization and temporal Refinement},
  author={Carl Doersch and Yi Yang and Mel Vecer{\'i}k and Dilara Gokay and Ankush Gupta and Yusuf Aytar and Jo{\~a}o Carreira and Andrew Zisserman},
  journal={2023 IEEE/CVF International Conference on Computer Vision (ICCV)},
  year={2023},
  pages={10027-10038},
}

@inproceedings{cotracker,
  title     = {CoTracker: It is Better to Track Together},
  author    = {Nikita Karaev and Ignacio Rocco and Benjamin Graham and Natalia Neverova and Andrea Vedaldi and Christian Rupprecht},
  booktitle = {Proc. {ECCV}},
  year      = {2024}
}

@inproceedings{taptr,
  title={{TAPTR: Tracking Any Point with Transformers as Detection}},
  author={Li, Hongyang and Zhang, Hao and Liu, Shilong and Zeng, Zhaoyang and Ren, Tianhe and Li, Feng and Zhang, Lei},
  booktitle={European Conference on Computer Vision},
  pages={57--75},
  year={2024},
}

@inproceedings{mlpmixer,
author = {Tolstikhin, Ilya and Houlsby, Neil and Kolesnikov, Alexander and Beyer, Lucas and Zhai, Xiaohua and Unterthiner, Thomas and Yung, Jessica and Steiner, Andreas and Keysers, Daniel and Uszkoreit, Jakob and Lucic, Mario and Dosovitskiy, Alexey},
title = {MLP-mixer: an all-MLP architecture for vision},
year = {2021},
booktitle = {Proceedings of the 35th International Conference on Neural Information Processing Systems},
articleno = {1857},
numpages = {12},
series = {NIPS '21}
}

@inproceedings{attetion,
  title={Attention is All you Need},
  author={Ashish Vaswani and Noam M. Shazeer and Niki Parmar and Jakob Uszkoreit and Llion Jones and Aidan N. Gomez and Lukasz Kaiser and Illia Polosukhin},
  booktitle={Neural Information Processing Systems},
  year={2017},
}

@article{MegaDepthLS,
  title={MegaDepth: Learning Single-View Depth Prediction from Internet Photos},
  author={Zhengqi Li and Noah Snavely},
  journal={2018 IEEE/CVF Conference on Computer Vision and Pattern Recognition},
  year={2018},
  pages={2041-2050},
}

@ARTICLE {midas,
    author  = "Ren\'{e} Ranftl and Katrin Lasinger and David Hafner and Konrad Schindler and Vladlen Koltun",
    title   = "Towards Robust Monocular Depth Estimation: Mixing Datasets for Zero-Shot Cross-Dataset Transfer",
    journal = "IEEE Transactions on Pattern Analysis and Machine Intelligence",
    year    = "2022",
    volume  = "44",
    number  = "3"
}

@inproceedings{selfsupervised,
  title={Unsupervised CNN for Single View Depth Estimation: Geometry to the Rescue},
  author={Ravi Garg and B. V. Kumar and G. Carneiro and Ian D. Reid},
  booktitle={European Conference on Computer Vision},
  year={2016},
  url={https://api.semanticscholar.org/CorpusID:299085}
}

@article{stablediffusion,
  title={High-Resolution Image Synthesis with Latent Diffusion Models},
  author={Robin Rombach and A. Blattmann and Dominik Lorenz and Patrick Esser and Bj{\"o}rn Ommer},
  journal={2022 IEEE/CVF Conference on Computer Vision and Pattern Recognition (CVPR)},
  year={2021},
  pages={10674-10685},
}

@article{dinov2,
  title={DINOv2: Learning Robust Visual Features without Supervision},
  author={Maxime Oquab and Timoth{\'e}e Darcet and Th{\'e}o Moutakanni and Huy Q. Vo and Marc Szafraniec and Vasil Khalidov and Pierre Fernandez and Daniel Haziza and Francisco Massa and Alaaeldin El-Nouby and Mahmoud Assran and Nicolas Ballas and Wojciech Galuba and Russ Howes and Po-Yao (Bernie) Huang and Shang-Wen Li and Ishan Misra and Michael G. Rabbat and Vasu Sharma and Gabriel Synnaeve and Huijiao Xu and Herv{\'e} J{\'e}gou and Julien Mairal and Patrick Labatut and Armand Joulin and Piotr Bojanowski},
  journal={ArXiv},
  year={2023},
  volume={abs/2304.07193},
}

@inproceedings{depthanything,
      title={Depth Anything: Unleashing the Power of Large-Scale Unlabeled Data}, 
      author={Yang, Lihe and Kang, Bingyi and Huang, Zilong and Xu, Xiaogang and Feng, Jiashi and Zhao, Hengshuang},
      booktitle={CVPR},
      year={2024}
}

@inproceedings{deberta,
title={DEBERTA: DECODING-ENHANCED BERT WITH DISENTANGLED ATTENTION},
author={Pengcheng He and Xiaodong Liu and Jianfeng Gao and Weizhu Chen},
booktitle={ICLR},
year={2021},
}

@inproceedings{nyudepth,
  author    = {Nathan Silberman, Derek Hoiem, Pushmeet Kohli and Rob Fergus},
  title     = {Indoor Segmentation and Support Inference from RGBD Images},
  booktitle = {ECCV},
  year      = {2012}
}

@inproceedings{SKYSCENES,
  title={SKYSCENES: A Synthetic Dataset forAerial Scene Understanding},
  author={ Khose, Sahil  and  Pal, Anisha  and  Agarwal, Aayushi  and Deepanshi and  Hoffman, Judy  and  Chattopadhyay, Prithvijit },
  booktitle={European Conference on Computer Vision},
  year={2025},
}

@inproceedings{adamw,
  title={Decoupled Weight Decay Regularization},
  author={Ilya Loshchilov and Frank Hutter},
  booktitle={International Conference on Learning Representations},
  year={2017},
}

@article{kitti,
  author = {Andreas Geiger and Philip Lenz and Christoph Stiller and Raquel Urtasun},
  title = {Vision meets Robotics: The KITTI Dataset},
  journal = {International Journal of Robotics Research (IJRR)},
  year = {2013}
}

@inproceedings{DepthCrafter,
            author      = {Hu, Wenbo and Gao, Xiangjun and Li, Xiaoyu and Zhao, Sijie and Cun, Xiaodong and Zhang, Yong and Quan, Long and Shan, Ying},
            title       = {DepthCrafter: Generating Consistent Long Depth Sequences for Open-world Videos},
            booktitle   = {CVPR},
            year        = {2025}
    }

@misc{chronodepth,
      title={Learning Temporally Consistent Video Depth from Video Diffusion Priors}, 
      author={Jiahao Shao and Yuanbo Yang and Hongyu Zhou and Youmin Zhang and Yujun Shen and Vitor Guizilini and Yue Wang and Matteo Poggi and Yiyi Liao},
      year={2024},
      eprint={2406.01493},
      archivePrefix={arXiv},
}

@misc{zoedepth,
  author = {Bhat, Shariq Farooq and Birkl, Reiner and Wofk, Diana and Wonka, Peter and Müller, Matthias},
  title = {ZoeDepth: Zero-shot Transfer by Combining Relative and Metric Depth},
  publisher = {arXiv},
  year = {2023},
}

@inproceedings{IEBins,
title={IEBins: Iterative Elastic Bins for Monocular Depth Estimation},
author={Shao, Shuwei and Pei, Zhongcai and Wu, Xingming and Liu, Zhong and Chen, Weihai and Li, Zhengguo},
booktitle={Advances in Neural Information Processing Systems (NeurIPS)},
year={2023}
}

@inproceedings{P3Depth,
  author    = {Patil, Vaishakh and Sakaridis, Christos and Liniger, Alex and Van Gool, Luc},
  title     = {P3Depth: Monocular Depth Estimation with a Piecewise Planarity Prior},
  booktitle = {Proceedings of the IEEE/CVF Conference on Computer Vision and Pattern Recognition (CVPR)},
  year      = {2022},
}

@inproceedings{diffusionsfm,
  title={DiffusionSfM: Predicting Structure and Motion via Ray Origin and Endpoint Diffusion}, 
  author={Qitao Zhao and Amy Lin and Jeff Tan and Jason Y. Zhang and Deva Ramanan and Shubham Tulsiani},
  booktitle={CVPR},
  year={2025}
}

@inproceedings{vibration_simu,
	address = {Chengdu City, China},
	title = {Simulation and {Analysis} of {Vibration} {Blurred} {Images}},
	isbn = {978-1-4244-3708-5},
	booktitle = {2010 {International} {Conference} on {Computational} {Intelligence} and {Software} {Engineering}},
	publisher = {IEEE},
	author = {Wang, Xiaoyan and Yi, Tang and Tang, Qiuyan and Feng, Liang and Ni, Guoqiang and Zhou, Liwei},
	month = sep,
	year = {2010},
	pages = {1--4},
}

@article{simu_res,
	title = {Imaging simulation and restoration for a mobile-based long-focus camera with optical stabilization},
	volume = {63},
	issn = {1559-128X, 2155-3165},
	number = {19},
	journal = {Applied Optics},
	author = {Qiu, Tianchen and Zhou, Jingwen and Feng, Huajun and Li, Qi and Li, Tongyue and Chen, Yueting},
	month = jul,
	year = {2024},
	note = {Publisher: Optica Publishing Group},
	pages = {F80},
}

@misc{seurat,
      title={Seurat: From Moving Points to Depth}, 
      author={Seokju Cho and Jiahui Huang and Seungryong Kim and Joon-Young Lee},
      year={2025},
      eprint={2504.14687},
      archivePrefix={arXiv},
      primaryClass={cs.CV},
}

@inproceedings{degradation,
  author = {Chen, Shiqi and Feng, Huajun and Gao, Keming and Xu, Zhihai and Chen, Yueting},
  title = {Extreme-Quality Computational Imaging via Degradation Framework},
  booktitle = {Proceedings of the IEEE/CVF International Conference on Computer Vision (ICCV)},
  month = oct,
  year = {2021},
  pages = {2632-2641},
}

@article{computational_optics,
  author = {Chen, Shiqi and Lin, Ting and Feng, Huajun and Xu, Zhihai and Li, Qi and Chen, Yueting},
  journal = {IEEE Transactions on Pattern Analysis and Machine Intelligence},
  title = {Computational Optics for Mobile Terminals in Mass Production},
  month = apr,
  year = {2023},
  volume = {45},
  number = {4},
  pages = {4245-4259},
}

\end{document}